\documentclass[conference]{IEEEtran}
\IEEEoverridecommandlockouts

\usepackage{cite}
\usepackage{amsmath,amssymb,amsfonts}
\usepackage{algorithm}
\usepackage{algorithmic}
\usepackage{graphicx}
\usepackage{textcomp}
\usepackage{booktabs}
\usepackage{hyperref}
\usepackage{xcolor}
\def\BibTeX{{\rm B\kern-.05em{\sc i\kern-.025em b}\kern-.08em
    T\kern-.1667em\lower.7ex\hbox{E}\kern-.125emX}}
\begin{document}

\title{RoGA: Towards Generalizable Deepfake Detection through Robust Gradient Alignment}
\author{Lingyu Qiu\textsuperscript{\rm 1,\rm 2}, Ke Jiang\textsuperscript{\rm 1,\rm 2}, Xiaoyang Tan\textsuperscript{\rm 1,\rm 2}$^{\dagger}$ \\

\textsuperscript{\rm 1}College of Computer Science and Technology,
Nanjing University of Aeronautics and Astronautics\\ \textsuperscript{\rm 2}MIIT Key Laboratory of Pattern Analysis and Machine Intelligence\\

\{qiulingyu,ke\_jiang,x.tan\}@nuaa.edu.cn\\
}
\maketitle
\renewcommand{\thefootnote}{}
\footnotetext{$^{\dagger}$ Corresponding author.}

\begin{abstract}
Recent advancements in domain generalization for deepfake detection have attracted significant attention, with previous methods often incorporating additional modules to prevent overfitting to domain-specific patterns. However, such regularization can hinder the optimization of the empirical risk minimization (ERM) objective, ultimately degrading model performance. In this paper, we propose a novel learning objective that aligns generalization gradient updates with ERM gradient updates. The key innovation is the application of perturbations to model parameters, aligning the ascending points across domains, which specifically enhances the robustness of deepfake detection models to domain shifts. This approach effectively preserves domain-invariant features while managing domain-specific characteristics, without introducing additional regularization. Experimental results on multiple challenging deepfake detection datasets demonstrate that our gradient alignment strategy outperforms state-of-the-art domain generalization techniques, confirming the efficacy of our method. The code is available at
\url{https://github.com/Lynn0925/RoGA}
.
\end{abstract}


\section{Introduction}
Highly realistic forgery face images generated by deep-learning methods, such as Deepfake \cite{lyu2020deepfake}, pose a huge threat to social security\cite{qiu2024multi}. Therefore, numerous detection methods \cite{RECCE, DFDC_Capsule, DFDC_Frequency3} have emerged as a defense technology with the help of deep networks to safeguard over the past time.

Although these previous deepfake detection methods have explored the traces left by forgery from various aspects, they often greatly suffer from poor generalization when exposed to unseen samples at the test stage, especially those generated by unknown manipulation methods, i.e., samples from different data domains. Such limitation significantly hinders the practical application of current deepfake detection methods. To address this issue, recent works\cite{choi2024exploiting} introduce prior assumptions over the domain, such as statistics of latent features, and extract the domain-invariant features to enhance the generalization over cross-domain manipulations through techniques like style-mixture \cite{mixup}. However, such assumptions are often too strong and hard to satisfy in practice\cite{zhang2020does}, which hinders these methods' generalization when deploying practically. Besides, additional regularization is often harmful to the convergence to the optimal solution of the empirical risk minimization (ERM) frameworks. These two points motivate us to develop a general and regularization-free method to deal with cross-domain deepfake detection.

\begin{figure}
     \centering
     \includegraphics[width=0.8\linewidth]{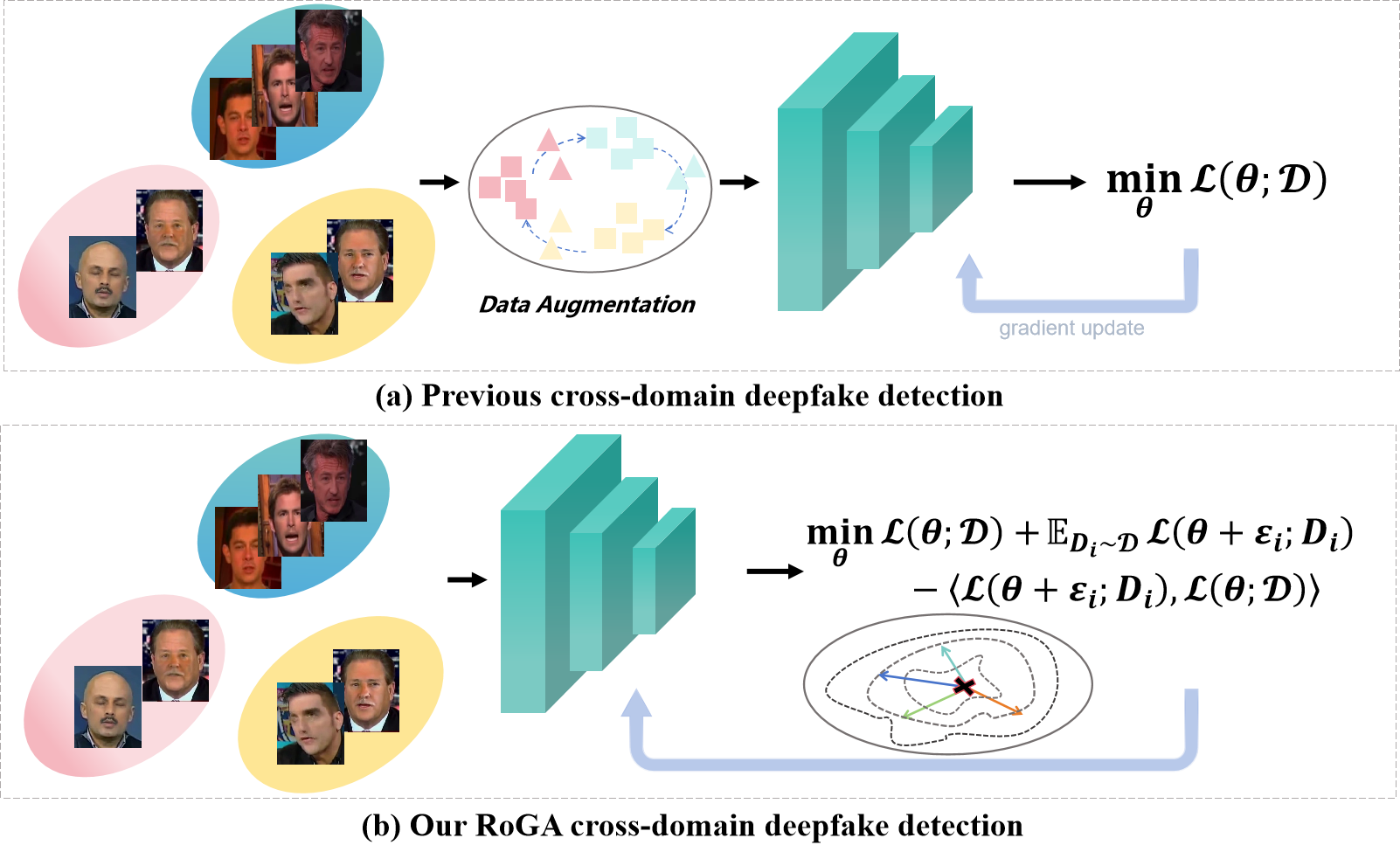}
     \caption{Comparison among previous cross-domain deepfake detection based on domain adaption such as data/feature augmentation. Our approach in (b) aligns the perturbed gradients, enhancing robustness and generalization.}
     \label{fig:enter-label}
 \end{figure}

In this work, we empirically attribute the challenges of cross-domain deepfake detection to the overfitting of neural network parameters to specific domains, resulting in difficulties in cross-domain generalization during the testing phase. This causes neural networks,
as illustrated in (a) of Fig.\ref{fig:enter-label}, 
learned by traditional methods to focus more on domain-specific features while neglecting broader patterns. Consequently, the learned models would perform well in specific domains, such as certain domains within the dataset, but exhibit poor performance in other domains, particularly when encountering data from unseen domains during testing.

To address this issue, we introduce a novel method in the domain generalization scenario of general face forgery detection, which we term as \textbf{Robust Gradient Alignment}  (RoGA) for deepfake detection. The key innovation of RoGA is to spontaneously apply perturbations to the model parameters and align the ascending points on each domain during the gradient updates, as is illustrated in (b) of Fig.\ref{fig:enter-label}. This unique approach guides deepfake detection models to be robust to deal with domain shift, as it can effectively preserve the domain-invariant features while appropriately handling domain-specific characteristics.
To be specific, this method attaches certain degrees of perturbation onto the learned models' parameters during the gradient descent process in the parameter space. This perturbation can help the model converge to a flatter local minimum, hence enjoying better robustness. In other words, this approach makes it more challenging for the learned model parameters to get trapped in the local optimal solution corresponding to a specific domain, thereby alleviating the overfitting phenomenon of the learned model to that particular domain.

Although our method shares similarities with the recent SAM\cite{sam} approach - both apply perturbations to model parameters to find a flatter minimum - there are key differences between our basic idea and SAM: 
(1) SAM does not consider the gradient alignment between different domains, which can influence the direction of the perturbation gradient during the descent step. 
(2) To our knowledge, parameter-robust methods like SAM have yet to be applied in the context of deepfake detection.

To further address the aforementioned issues, 
we introduce a novel objective in domain generalization scenario in deepfake detection which is designed to guide the detectors towards an optimal flat minimum towards robust domain shift.
The objective can be divided into two parts:
1) the objective should aim for an optimal low minima under perturbation during the training process.
2) Considering the perturb of different domain's alignment, the gradient updates for each domain should be aligned with each other.
This dual objective enables our model to learn a more stable local minimum and more robust domain shifts over different face forgery datasets.
All in all, the main contributions of our work are concluded as follows:
\begin{itemize}
    \item We provide a novel perspective on cross-domain deepfake detection. Rather than designing distinct representation learning approaches to extract domain-invariant features, as is common in previous work, we enhance generalization and domain transferability by guiding the model toward the optimal flat and robust minimum by adding perturb during gradient updates.

    \item We propose a new training objective:
During gradient updates at the ascending point, we keep the generalization gradients of each domain consistent with the empirical risk minimum gradient update, which is conducive to domain generalization in deepfake detection.

    \item We demonstrate that our approach outperforms well-established baselines in performance across a range of popular deepfake detection evaluation protocol settings.
\end{itemize}

\section{Related Work}

In deepfake detection, domain generalization methods based on invariant representation learning\cite{CORE} have been widely used. A common method is through data augmentation\cite{luo2021generalizing,UCF,yang2024adaforensics}, such as SLADD\cite{sladd} dynamically synthesizing data through adversarial methods, while SBI\cite{sbi} swapping with the identity of the same person. On the other hand, there are also studies that solve the new problem of general face forgery detection through meta-learning\cite{sun2021domain,MLDG} and multi-task settings\cite{Multi-task}. RECCE\cite{RECCE} learns realistic face representations by incorporating self-supervised modules. Previous work, which often uses auxiliary modules\cite{DFDC_Frequency3,WildDeepfake} to remove domain-specific features, struggles to generalize to unseen domains due to uncertainties in achieving flat loss regions during training.

As for the optimization strategy of domain gradient consistency, ConfR \cite{chen2024confr} introduces a novel learning objective to reduce gradient conflicts between domains and extract common features across forgeries, but it overlooks the robust generalization to unknown domains.
In contrast, our approach leverages the inherent sharpness of a model within specific domains by aligning these models. In summary, our goal is to construct a more robust and universally generalizable model for deepfake detection.

\begin{figure*}
    \centering
    \includegraphics[width=1\linewidth]{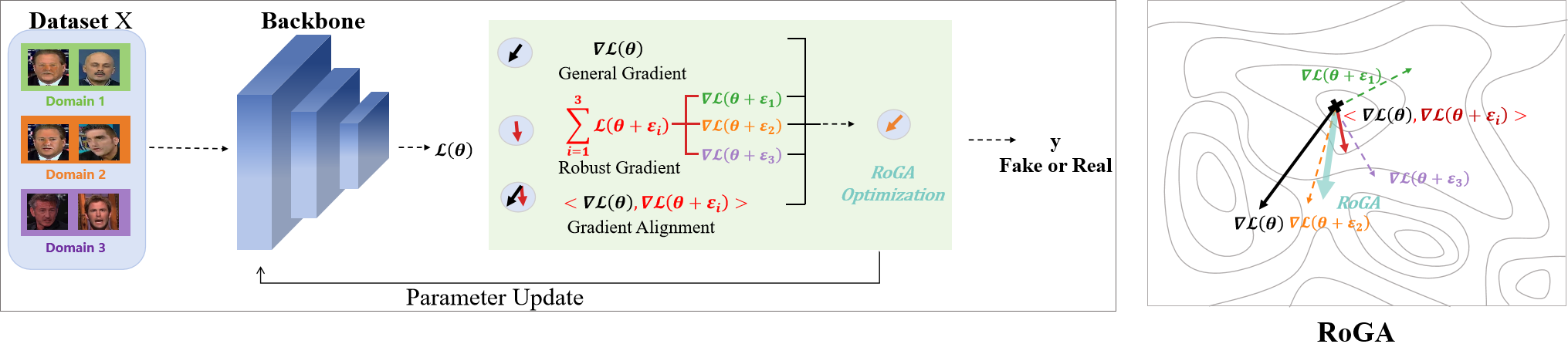}
    \caption{The left part illustrates the framework of RoGA. The right part demonstrates the basic idea of RoGA, where it aligns the gradients with different perturbations during training, hence reducing the risk of overfitting to specific domain patterns.}
    \label{fig:RoDA}
\end{figure*}

\section{Preliminaries}
A face forgery detection problem could be formulated as a binary classification problem. 
To be specific, the observed data $(x, y)$ are assumed to be sampled from a fixed but unknown joint distribution $P(X, Y)$. 
To model this relationship, we assume that there exists a neural network $f : X \to Y$, whose weight parameters are represented by $\theta$. Its purpose is to learn a set of parameters $\hat{\theta}$ through the source dataset $D$ to determine whether the input $x$ is \textit{real} or \textit{fake}. Let $\mathcal{L}$ be some loss function, then the standard
learning objective of empirical risk minimization can be defined as follows:
\begin{align}
    \min_{\theta} \mathbb E_{(x,y)\sim P(X,Y)} [\mathcal L(f(x;\theta), y)]\label{pre:objective}
\end{align}

To generalize this formulation to the cross-domain setting, suppose there exists a domain distribution $P(\mathcal{D})$, then we can sample a set of domain $d$ on it, i.e., $D\sim P(\mathcal{D})$.

According to such understanding of domain information, the data distribution could be divided by different domains, i.e., $P(X,Y) = \mathbb E_{D\sim P(\mathcal{D})}P(X,Y|D)$. Traditional methods \cite{chen2024confr,MLDG,Multi-task} formulate the objective of cross-domain face forgery detection in an expected way, i.e.,
\begin{align}
    \min_{\theta} \mathbb E_{D\sim P(\mathcal{D})}\mathbb E_{(x,y)\sim P(X,Y|D)} [\mathcal L(f(x;\theta), y)]\label{eq:expected_obj}
\end{align}

The ERM objective function optimizes average performance across all domains but may lead to overfitting on shortcut features and convergence to a sharp minimum, impairing generalization to other domains. Inspired by SAM\cite{sam}, we introduce perturbations during the parameter update process to guide the model toward a flatter minimum, enhancing its robustness, as demonstrated in previous studies. 

\section{Robustness Gradient Alignment}
In this section, we propose a method named Robustness Gradient Alignment (RoGA). Specifically, we first introduce the expected perturbation optimization on ERM inspired by SAM\cite{sam}; the second part acts as a conservative term to ensure that the perturbation in the optimization direction will not differ too much from the original one.

Subsequently, we analyze it and propose an implementable algorithm.
\subsection{Robustness Gradient Optimization}

For a single domain, we use the following SAM\cite{sam} objective:
\begin{align}
    \min_\theta \max_{\|\epsilon_i\|\leq\rho} \mathcal{L}(\theta+\epsilon_i; \mathcal{D}_{i})\label{eq:perturb_loss}
\end{align}
where $\epsilon_i$ denotes some perturbation imposed on the parameter $\theta$ for the $i-$th domain. The objective penalizes the model by its 'sharpness' or sensitivity  to the small perturbation $\epsilon_i$ at $\theta$. Using the Taylor expansion around $\epsilon_i$, we can transform the inner maximization in Eq. \ref{eq:perturb_loss} into a linearly constrained optimization, yielding the following solution:

\begin{align}
\epsilon^{*}_{i}=arg\max_{\|\epsilon\|\leq\rho} \mathcal{L}(\theta+\epsilon; \mathcal{D}_{i}) \approx \rho\frac{\nabla\mathcal L(\theta; \mathcal{D}_{i})}{\|\nabla\mathcal L(\theta; \mathcal{D}_{i})\|}
\end{align}

With this, we construct the following objective for our cross-domain deepfake detection, which basically seeks to minimize the empirical robust loss function over multiple ($K$) domains,

\begin{align}
\min_\theta  \frac{1}{K}\sum_{i=1}^{K}\mathcal{L}(\theta + \hat{\epsilon}_i; \mathcal{D}_{i})
\text{ where }  \hat{\epsilon}_{i}\stackrel{\triangle }{=} \rho\frac{\nabla\mathcal L(\theta; \mathcal{D}_{i})}{\|\nabla\mathcal L(\theta; \mathcal{D}_{i})\| }\label{eq:multi_SAMloss}
\end{align}

It is noteworthy that there exists an important difference between the above objective function (Eq.~\ref{eq:multi_SAMloss}) and the original (Eq.~\ref{eq:expected_obj}). That is,  the original objective was a “strong requirement,” which means that the model was expected to simultaneously interpret data from multiple domains. In reality, achieving this goal is quite challenging and is prone to be over-fitting. However, our new objective function (Eq.~\ref{eq:multi_SAMloss}) relaxes this requirement. It merely demands that the model learn an optimal shared position in the parameter space, allowing it to interpret data from different domains after a single-step gradient adjustment. This relaxation is more reasonable than the original strong-fitting criterion.

\subsection{Domain Aware Gradient Alignment}

To further improve the performance, we consider the issue due to gradient misalignment caused by domain artifacts and data imbalance in multidomain settings. As Fig. \ref{fig:RoDA} shows, domain-specific perturbations (e.g., $\nabla\mathcal{L}(\theta+\epsilon_1;\mathcal{D}_1)$) may significantly deviate from the main gradient direction $\nabla\mathcal{L}(\theta;\mathcal{D})$, compromising optimization stability. 

To address this issue, we require that the perturbed gradient $\nabla\mathcal{L}(\theta +\epsilon_i; \mathcal{D}_i)$ for each domain remain aligned with its empirical risk gradient $\nabla\mathcal{L}(\theta; \mathcal{D}_i)$. Finally, we obtain the following Robustness Gradient Alignment (RoGA) optimization loss:
\begin{align}
 \frac{1}{K}\sum_{i=1}^{K}[\mathcal{L}(\theta+\epsilon_i; \mathcal{D}_{i})
 -\alpha\langle \nabla\mathcal{L}(\theta+\epsilon_i; \mathcal{D}_{i}),\nabla\mathcal{L}(\theta; \mathcal{D}_i)\rangle]\label{eq:loss_ROGA}
\end{align}
where $\alpha$ is the balance coefficient. Note that this is a general objective - as shown later, it can be easily incorporated into any cross-domain fake face detection algorithms. During the optimization process, we decoupled $\epsilon$ from the parameters $\theta$ by first estimating the $\epsilon$ value with the current $\theta_t$ and then update the $\theta_{t+1}$ value by stochastic gradient decent.

\section{Experimental Result}
\subsection{Experiment Settings}
\vspace{-3pt}
\textbf{ Datasets.}
 To evaluate the generalization of our method, we conduct experiments on widely used deepfake datasets: FaceForensics++ (FF++)\cite{FF++}, Deepfake Detection Challenge (DFDC)\cite{DFDC}, CelebDF (v1, v2)\cite{Celeb-DF}, DFDCP\cite{dolhansky2019dee}, and UADFV\cite{liy2018exposingaicreated}. FF++ includes four manipulation methods—DeepFakes (DF), Face2Face (F2F), FaceSwap (FS), and NeuralTexture (NT)—with two compression levels: low quality (c23) and high quality (c40).

\textbf{Training Details.}
ResNet34\cite{CNN-Aug} serves as the baseline architecture for feature learning, with parameters initialized via ImageNet pre-training\cite{imagenet}. The model is optimized using SGD\cite{sgd} as the base optimizer with a learning rate of 0.005.
 The hyperparameter $\alpha$ is set as 0.0002, while  $\rho$ = 0.1.

\textbf{Evaluation Metrics.}
For a comprehensive evaluation, we evaluate the performance of the detector at image-level and report Area Under Curve (AUC) and Accuracy (ACC) as primary metrics. Average Precision (AP) and Equal Error Rate (EER) are also included for additional insights. 
We ensure a fair comparison by adhering to the experimental settings defined by DeepfakeBench\cite{DeepfakeBench}.

\begin{table*}
\centering
\caption{ Comparative performance for various domain-adaption-based methods under cross-dataset evaluation using AUC(\%) values. We divide the methods into four categories, "Naive" represents without use of domain generalization, and the others represent the categories of domain generalization. Then, we bold the \textbf{Top.1} methods with the best cross-domain performance.}
\begin{tabular}{l|c|c|ccccc}
\toprule
Method & Detector & Backbone & CDF-v1 & CDF-v2 & UADFV & DFDC & DFDCP \\
\midrule
Naive & Meso4 \cite{meso4} & MesoNet & 0.736 & 0.609 &0.715 & 0.556 & 0.599 \\
Naive & CNN-Aug\cite{CNN-Aug} & ResNet & 0.742 & 0.703 & 0.874 & 0.636 & 0.617 \\
Naive & Xception\cite{Xception}& Xception & 0.779 & 0.737 & 0.937 & 0.708 & 0.737 \\
Naive & EfficientB4\cite{tan2019efficientnet} & EfficientNet & 0.791 & 0.749 & 0.947 & 0.696 & 0.728 \\
\hline
Representation Learning & SPSL\cite{DFDC_Frequency3} & Xception & 0.815 & 0.765 & 0.942 & 0.704 & 0.741 \\
Representation Learning& CORE\cite{CORE} & Xception & 0.780 & 0.743 & 0.941 & 0.705 & 0.734  \\
Representation Learning& Recce\cite{RECCE}& Designed & 0.768 & 0.732 & 0.945 & 0.713 & 0.734  \\

\hline

Data Augmentation & SRM\cite{luo2021generalizing} & Xception & 0.793 & 0.755 & 0.942 & 0.700 & 0.741\\
Data Augmentation & UCF\cite{UCF} & Xception & 0.779 & 0.753 & 0.953 & 0.719 & 0.759  \\
Data Augmentation& AdaForensics\cite{yang2024adaforensics}&ResNet&0.869& 0.793&0.955 & 0.747 &0.779 \\
\hline
Learning Strategy & MLDG\cite{MLDG} & Xception & 0.641 & - & -&0.682& -\\
Learning Strategy & Multi-task\cite{Multi-task} & Xception &0.609 & -& - &0.682& -\\
Learning Strategy & LTW\cite{sun2021domain} & Xception &0.609 & - & -&0.690& - \\
Learning Strategy &ConfR\cite{chen2024confr}& EfficientNet & 0.873&- &-& \textbf{0.803}&\textbf{0.828}\\
\hline
Ours(RoGA) & ResNet34 & ResNet & \textbf{0.877}& \textbf{0.858} &\textbf{0.959} & 0.751 &  0.753 \\
 
 \bottomrule
 \end{tabular}
    \label{cross-dataset-sota}
\end{table*}
\vspace{-0.2cm}
\subsection{Domain Generalization Evaluation}
\textbf{Cross-Datasets Evaluation}
To assess the generalization capacity of our method, we follow a standard cross-dataset evaluation protocol. The models are trained on FF++(c23)\cite{FF++} and tested on unseen datasets, including Celeb-DF\cite{Celeb-DF} and DFDC\cite{DFDC}, serving as target domains to evaluate domain robustness.
Table \ref{cross-dataset-sota} shows that our method outperforms state-of-the-art approaches, achieving the highest AUC of 0.877 on Celeb-DF and 0.751 on DFDC. The results highlight the efficacy of our multi-source gradient alignment strategy, which enhances domain-agnostic feature learning without introducing additional complexity. Compared to domain adaptation and data augmentation-based methods, our approach demonstrates superior robustness across unseen datasets.

\textbf{Muli-source Cross-Manipulation Evaluation}
To verify the effectiveness of the proposed Domain-Aware Robustness Gradient Alignment(RoGA), we conducted experiments on the multi-source forgery method of FF++\cite{FF++}. Specifically, models are trained on three forgery methods while incorporating domain labels and tested on the remaining forgery method as the target domain. This setup simulates diverse manipulation transformations originating from identical source images, thus providing an evaluation of cross-manipulation generalization.
Table \ref{cross_manupulation_excluded1} reports the experimental results consistently outperform competitive baselines, particularly achieving remarkable AUC of 98.08\% for GID-FS and 90.08\% for GID-DF, which are \textbf{11.78\%} and \textbf{3.98\%} higher than the second detector, respectively.
These results demonstrate the effectiveness of gradient alignment in capturing subtle manipulative variations, significantly boosting the model's generalization to unseen forgery methods. In summary, our approach introduces a robust optimization strategy, advancing cross-domain deepfake detection with superior performance across diverse datasets and forgery techniques.

\begin{table}[htbp]
    \caption{Multi-source evaluation results on ACC(\%)/AUC (\%)."GID-DF" means training on the FF++ excluded DF while testing in DF.}
    \centering
    \begin{tabular}{l|ccccc}
        \toprule
         Methods& GID-DF & GID-F2F & GID-FS & GID-NT  \\
        \midrule
        MLDG\cite{MLDG}& 67.2/73.1 &58.1/61.7 &58.1/61.7 &56.9/60.7 \\
        LTW\cite{sun2021domain}&  69.1/75.6 &65.7/72.4 &62.5/68.1 &58.5/60.8 \\
        DisGRL\cite{shi2023discrepancy}& 77.3/86.1 &75.8/84.3 &76.9/86.3 &\textbf{66.3}/\textbf{72.8} \\
        Ours(RoGA)&  \textbf{80.13}/\textbf{90.08} &     \textbf{76.34}/\textbf{86.54}
         &           \textbf{93.41}/\textbf{98.08}&59.43/70.52\\
        \bottomrule
    \end{tabular}

    \label{cross_manupulation_excluded1}
\end{table}
\vspace{-4pt}
\subsection{Ablation Study}
\subsubsection{Effects of proposed objectives}
Next, we conducted experiments on the FF++(c23) benchmark and selected ResNet34\cite{CNN-Aug} as the backbone to investigate the effectiveness of the two objectives of \textbf{RoGA}.
Specifically, (a) denotes the baseline optimizer (SGD\cite{sgd}), (b) incorporates the robust gradient term $\frac{1}{K}\sum_{i=1}^{K}\mathcal{L}(\theta+\epsilon_i; \mathcal{D}_{i})$, and (c) adds the domain-aware alignment term $\langle \nabla\mathcal{L}(\theta+\epsilon_i; \mathcal{D}_{i}),\nabla\mathcal{L}(\theta; \mathcal{D}_i)\rangle$, with (d) representing the complete method, it achieve an optimal AUC performance of \textbf{99.30\%}, surpassing individual gains of 0.27\% and 0.10\%.
\vspace{-0.4cm}
\begin{table}[htbp]
\scriptsize
\centering
\caption{Ablation study: Analysis of the two objectives while training on the FF++(c23) and testing on the DeepFakes.}
\begin{tabular}{l|cccc}
\toprule
   & AUC $\uparrow$ & ACC $\uparrow$ & Loss $\downarrow$ & EER $\downarrow$ \\
\midrule
(a) Base-optimizer &97.65&90.00&0.32&7.83\\
(b) + $\mathcal{L}(\theta + \hat{\epsilon}_i; \mathcal{D}_{i})$ &99.03&93.39&0.18&4.79 \\
(c) + $\langle \nabla\mathcal{L}(\theta+\epsilon_i; \mathcal{D}_{i}),\nabla\mathcal{L}(\theta; \mathcal{D}_i)\rangle$ & 99.20&91.61&0.22&5.38\\
(d) RoGA & 99.30&94.25&0.15&4.13 \\
\bottomrule
\end{tabular}
\label{table:ablation_objective}
\end{table}
\subsubsection{Effects of the optimizer selection}
To assess the efficacy of our proposed optimizer, we conduct a comparative analysis with various optimization techniques across both cross-manipulation and cross-dataset settings. As shown in Table \ref{table:ablation_sam}, sharpness-aware optimizers like SAM\cite{sam} and ASAM\cite{asam} outperform traditional ERM-based optimizers such as Adam\cite{kingma2014adam} and SGD\cite{sgd}. Our RoGA optimizer, which integrates perturbation regularization and domain-specific gradient alignment, consistently delivers superior performance with a 1.47\% ACC improvement over Adam\cite{kingma2014adam} in-domain and a notable 6.1\% gain in cross-domain tests.
\begin{table}[htbp]
\caption{Ablation study: Alternative approach using domain gradient for Domain Generation under cross-manipulation and cross-dataset evaluation}
\centering
\begin{tabular}{c|cc|cc}
\toprule
Train& \multicolumn{2}{c|}{FS, F2F, NT} &\multicolumn{2}{c}{FF++}\\
Test& \multicolumn{2}{c|}{DF} &\multicolumn{2}{c}{UADFV}\\
\hline
 Optimizer  & AUC $\uparrow$ & ACC $\uparrow$ & AUC $\uparrow$ & ACC $\uparrow$ \\
\midrule
Adam\cite{kingma2014adam} & 92.96& 83.83&81.63 &55.98 \\
SGD \cite{sgd} & 92.27&84.82& 76.45&56.18  \\
SAM \cite{sam} & 93.18&83.39& 81.50&50.63   \\
SAGM \cite{SAGM} & 93.05&84.46&85.13&59.86 \\
Ours(RoGA) &   \textbf{93.63}&\textbf{85.30}&\textbf{95.95}&\textbf{62.08}  \\
\bottomrule
\end{tabular}
\label{table:ablation_sam}
\end{table}

\subsubsection{Effects of the backbone}\label{sec:ablation_bone}

We evaluate the flexibility of our method by testing it with various backbones, including Xception\cite{Xception}, EfficientNetB4\cite{tan2019efficientnet}, and ResNet34\cite{CNN-Aug}, which are commonly used in deepfake detection. Experiments were conducted on FF++(c23), and evaluated on four manipulation methods from FF++(c23) and cross-domain datasets UADFV\cite{liy2018exposingaicreated}. The results presented in Table \ref{table:ablation_backbone} (AUC)
unequivocally demonstrate that our method enhances performance without adding any additional model parameters, regardless of the backbone type. Specifically, we observe a \textbf{7.21\%} average performance increase on the Xception\cite{Xception} framework. This underscores the remarkable versatility of our approach, which exhibits seamless integration capabilities, thereby offering substantial performance gains across a diverse set of well-established detection frameworks. 

\vspace{-0.4cm}
\begin{table}[htbp]
\caption{Ablation study: The AUC(\%) values of alternative approach using \textbf{Ours RoGA} in Different Backbone}
\centering
\begin{tabular}{l|c|cccc|c}
\toprule
Backbone &+Ours& {DF}&{F2F}&{FS} &NT&UADFV\\
\midrule
Meso4\cite{meso4} & $\times$&75.89&70.38&61.31&66.95&64.49\\
 & $\checkmark$&79.84&75.57&64.57&67.61&71.78\\

Xception\cite{Xception} &$\times$&91.51&91.86&88.64&82.22&88.27\\
& $\checkmark$&99.00&99.08&99.06&96.42&91.55\\

 EffcientNet\cite{tan2019efficientnet} & $\times$&90.51&88.05&86.98&81.14&83.40\\
 & $\checkmark$&97.36&97.57&97.53&94.82&87.87\\
 ResNet34\cite{CNN-Aug} & $\times$&97.65&97.62&97.81&96.19&85.31\\
 &$\checkmark$&99.20&98.85&99.05&96.41&95.95\\
\bottomrule
\end{tabular}
\label{table:ablation_backbone}
\end{table}
\vspace{-4pt}
\subsubsection{Hyparameter Sensitivity}
In our proposed RoGA, the hyperparameters $\alpha$ and $\rho$ are critical for performance. Table \ref{table:ablation—hyper} presents the AUC(\%) results across intra- and cross-dataset settings on FF++\cite{FF++}. The optimal values, $\alpha$ = 0.001 and $\rho$ = 0.1, consistently yield superior results under both evaluations.
\vspace{-4pt}
\begin{table}[htbp]
\centering
\caption{Hyperparameters Sensitivity of $\alpha$ and $\rho$ under both intra- and cross-dataset settings where model is trained on FF++(c23)}
\begin{tabular}{cc|cccccccccccc}
\toprule
\multicolumn{2}{c}{Hyperparam.}&DF&F2F& FS &NT &UADFV&CelebDF\\
$\alpha$&$\rho$\\
\midrule
0.001&0.05&99.20&98.85&99.05&96.41&91.01&81.09\\
 0.002&0.05&99.02&98.75&99.12&95.88&90.44&80.61\\
  0.001&0.1&99.09&98.87&99.39&96.26&94.23&87.70\\

\bottomrule
\end{tabular}
\label{table:ablation—hyper}
\end{table}
\vspace{-4pt}

\vspace{-0.55cm}
\subsection{Analysis}
To enhance our understanding of the RoGA method, we use interpretable methods such as Grad-CAM\cite{GRADCAM} and t-SNE\cite{tsne} for analysis.

\begin{figure}
    \centering
    \includegraphics[width=\linewidth]{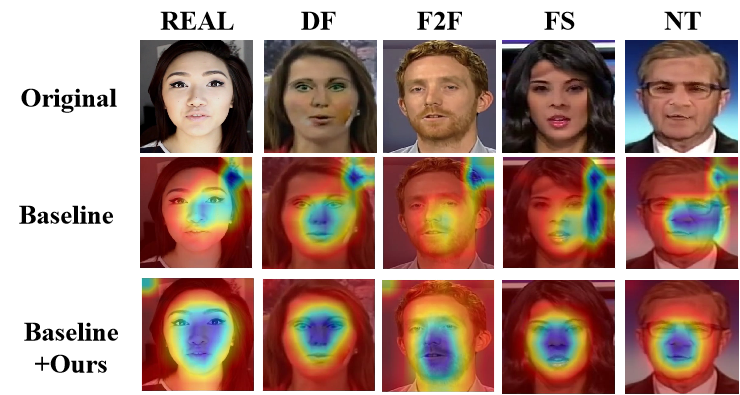}
    \caption{The GradCAM visualizations\cite{GRADCAM} comparing SRM \cite{super-resolution} baseline and our \textbf{RoGA}, across four forgery types on FF++(c23).}
    \label{fig:grad-cam}
\end{figure}

\textbf{Visualizations of the captured artifacts. }
GradCAM\cite{GRADCAM} is employed to visualize the regions most critical for the detector, offering interpretability of the model. As shown in Figure \ref{fig:grad-cam}, our RoGA significantly outperforms the SRM baseline\cite{super-resolution}, which focuses on uniform but limited artifact patterns. In contrast, \textbf{RoGA} precisely localizes manipulated facial regions, demonstrating its capability to learn robust and domain-invariant features.

\begin{figure}[htbp]
    \centering
    \includegraphics[width=\linewidth]{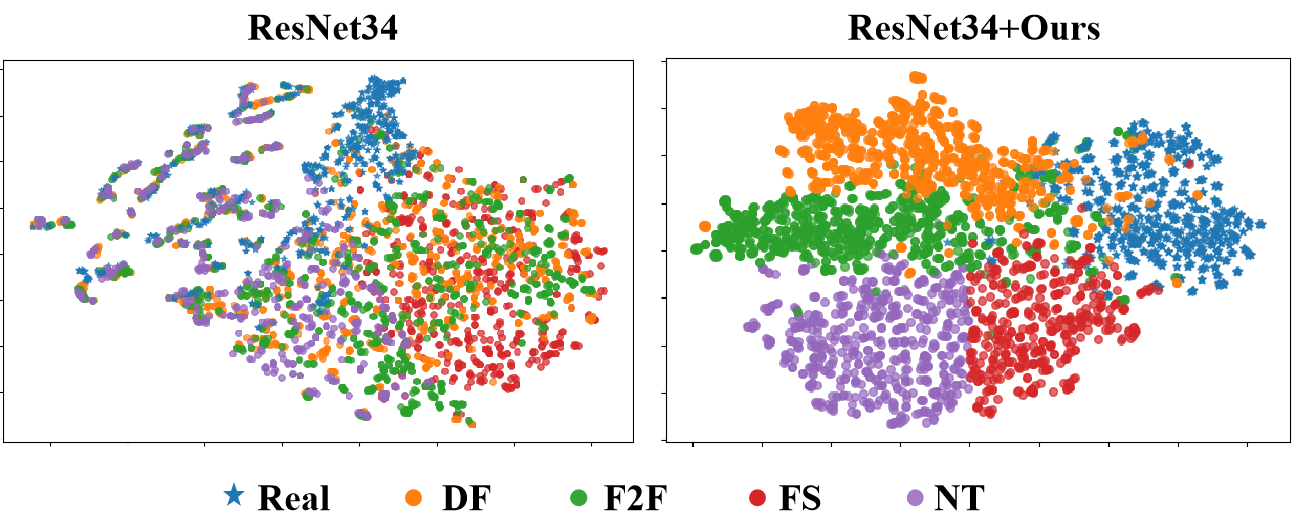}
    \caption{t-SNE\cite{tsne} visualization of latent space \textit{w} and \textit{w/o} our \textbf{RoGA} when the model is trained on FF++(c23).}
    \label{fig:t_sne}
\end{figure}

\textbf{Visualizations of learned latent space.}
\vspace{-2pt}
To assess the learned representations, we visualized the latent space of 5,000 test samples from FF++(c23) using t-SNE\cite{tsne}. As shown in Figure \ref{fig:t_sne}, our RoGA method indicates that our enhanced method (right) has indeed learned a much clearer and more distinct decision boundary compared to the unenhanced baseline (left).
\vspace{-0.55cm}
\vspace{-0.55cm}
\section{Conclusion}
We propose RoGA (Robustness Gradient Alignment), a novel optimization framework for domain generalization in deepfake detection. RoGA combines gradient perturbation and alignment, with its core innovation being gradient orthogonal alignment in the loss function. This ensures that perturbation and ERM gradients are harmonized, guiding updates towards robust, generalizable minima.

By modeling domain variability via perturbations and resolving inter-domain conflicts through alignment, RoGA effectively enhances cross-domain generalization. Extensive experiments both ablation studies and rigorous comparisons with state-of-the-art methods demonstrate RoGA’s adaptability and state-of-the-art performance across diverse architectures, underscoring its significance in advancing optimization strategies for multi-domain deepfake detection.

\textbf{Acknowledgement }This work is partially supported by the National Natural Science Foundation of China (6247072715).
\newpage
\bibliographystyle{IEEEbib}
\bibliography{icme2025}

\begin{thebibliography}{10}

\bibitem{lyu2020deepfake}
Siwei Lyu,
\newblock ``Deepfake detection: Current challenges and next steps,''
\newblock in {\em 2020 IEEE international conference on multimedia \& expo
  workshops (ICMEW)}. IEEE, 2020, pp. 1--6.

\bibitem{qiu2024multi}
Lingyu Qiu, Ke~Jiang, Sinan Liu, and Xiaoyang Tan,
\newblock ``Multi-level distributional discrepancy enhancement for cross domain
  face forgery detection,''
\newblock in {\em Chinese Conference on Pattern Recognition and Computer Vision
  (PRCV)}. Springer, 2024, pp. 508--522.

\bibitem{RECCE}
Junyi Cao, Chao Ma, Taiping Yao, Shen Chen, Shouhong Ding, and Xiaokang Yang,
\newblock ``End-to-end reconstruction-classification learning for face forgery
  detection,''
\newblock in {\em Proceedings of the IEEE/CVF Conference on Computer Vision and
  Pattern Recognition}, 2022, pp. 4113--4122.

\bibitem{DFDC_Capsule}
HuyH. Nguyen, Junichi Yamagishi, and Isao Echizen,
\newblock ``Capsule-forensics: Using capsule networks to detect forged images
  and videos,''
\newblock {\em Cornell University - arXiv,Cornell University - arXiv}, Oct
  2018.

\bibitem{DFDC_Frequency3}
Honggu Liu, Xiaodan Li, Wenbo Zhou, Yuefeng Chen, Yuan He, Hui Xue, Weiming
  Zhang, and Nenghai Yu,
\newblock ``Spatial-phase shallow learning: rethinking face forgery detection
  in frequency domain,''
\newblock in {\em Proceedings of the IEEE/CVF conference on computer vision and
  pattern recognition}, 2021, pp. 772--781.

\bibitem{choi2024exploiting}
Jongwook Choi, Taehoon Kim, Yonghyun Jeong, Seungryul Baek, and Jongwon Choi,
\newblock ``Exploiting style latent flows for generalizing deepfake video
  detection,''
\newblock in {\em Proceedings of the IEEE/CVF Conference on Computer Vision and
  Pattern Recognition}, 2024, pp. 1133--1143.

\bibitem{mixup}
Hongyi Zhang, Moustapha Cisse, Yann~N Dauphin, and David Lopez-Paz,
\newblock ``mixup: Beyond empirical risk minimization,''
\newblock {\em arXiv preprint arXiv:1710.09412}, 2017.

\bibitem{zhang2020does}
Linjun Zhang, Zhun Deng, Kenji Kawaguchi, Amirata Ghorbani, and James Zou,
\newblock ``How does mixup help with robustness and generalization?,''
\newblock {\em arXiv preprint arXiv:2010.04819}, 2020.

\bibitem{sam}
Pierre Foret, Ariel Kleiner, Hossein Mobahi, and Behnam Neyshabur,
\newblock ``Sharpness-aware minimization for efficiently improving
  generalization,''
\newblock {\em arXiv preprint arXiv:2010.01412}, 2020.

\bibitem{CORE}
Yunsheng Ni, Depu Meng, Changqian Yu, Chengbin Quan, Dongchun Ren, and Youjian
  Zhao,
\newblock ``Core: Consistent representation learning for face forgery
  detection,''
\newblock in {\em Proceedings of the IEEE/CVF conference on computer vision and
  pattern recognition}, 2022, pp. 12--21.

\bibitem{luo2021generalizing}
Yuchen Luo, Yong Zhang, Junchi Yan, and Wei Liu,
\newblock ``Generalizing face forgery detection with high-frequency features,''
\newblock in {\em Proceedings of the IEEE/CVF conference on computer vision and
  pattern recognition}, 2021, pp. 16317--16326.

\bibitem{UCF}
Zhiyuan Yan, Yong Zhang, Yanbo Fan, and Baoyuan Wu,
\newblock ``Ucf: Uncovering common features for generalizable deepfake
  detection,''
\newblock in {\em Proceedings of the IEEE/CVF International Conference on
  Computer Vision}, 2023, pp. 22412--22423.

\bibitem{yang2024adaforensics}
Xiaoke Yang, Haixu Song, Xiangyu Lu, Shao-Lun Huang, and Yueqi Duan,
\newblock ``Adaforensics: Learning a characteristic-aware adaptive deepfake
  detector,''
\newblock in {\em 2024 IEEE International Conference on Multimedia and Expo
  (ICME)}. IEEE, 2024, pp. 1--6.

\bibitem{sladd}
Liang Chen, Yong Zhang, Yibing Song, Lingqiao Liu, and Jue Wang,
\newblock ``Self-supervised learning of adversarial example: Towards good
  generalizations for deepfake detection,''
\newblock in {\em Proceedings of the IEEE/CVF conference on computer vision and
  pattern recognition}, 2022, pp. 18710--18719.

\bibitem{sbi}
Kaede Shiohara and Toshihiko Yamasaki,
\newblock ``Detecting deepfakes with self-blended images,''
\newblock in {\em Proceedings of the IEEE/CVF Conference on Computer Vision and
  Pattern Recognition}, 2022, pp. 18720--18729.

\bibitem{sun2021domain}
Ke~Sun, Hong Liu, Qixiang Ye, Yue Gao, Jianzhuang Liu, Ling Shao, and Rongrong
  Ji,
\newblock ``Domain general face forgery detection by learning to weight,''
\newblock in {\em Proceedings of the AAAI conference on artificial
  intelligence}, 2021, vol.~35, pp. 2638--2646.

\bibitem{MLDG}
Da~Li, Yongxin Yang, Yi-Zhe Song, and Timothy Hospedales,
\newblock ``Learning to generalize: Meta-learning for domain generalization,''
\newblock in {\em Proceedings of the AAAI conference on artificial
  intelligence}, 2018, vol.~32.

\bibitem{Multi-task}
Huy~H Nguyen, Fuming Fang, Junichi Yamagishi, and Isao Echizen,
\newblock ``Multi-task learning for detecting and segmenting manipulated facial
  images and videos,''
\newblock in {\em 2019 IEEE 10th international conference on biometrics theory,
  applications and systems (BTAS)}. IEEE, 2019, pp. 1--8.

\bibitem{WildDeepfake}
Bojia Zi, Minghao Chang, Jingjing Chen, Xingjun Ma, and Yu-Gang Jiang,
\newblock ``Wilddeepfake: A challenging real-world dataset for deepfake
  detection,''
\newblock in {\em Proceedings of the 28th ACM International Conference on
  Multimedia}, Oct 2020.

\bibitem{chen2024confr}
Jin Chen, Jiahe Tian, Cai Yu, Xi~Wang, Zhaoxing Li, Yesheng Chai, Jiao Dai, and
  Jizhong Han,
\newblock ``Confr: Conflict resolving for generalizable deepfake detection,''
\newblock in {\em 2024 IEEE International Conference on Multimedia and Expo
  (ICME)}. IEEE, 2024, pp. 1--6.

\bibitem{FF++}
Andreas Rossler, Davide Cozzolino, Luisa Verdoliva, Christian Riess, Justus
  Thies, and Matthias Niessner,
\newblock ``Faceforensics++: Learning to detect manipulated facial images,''
\newblock in {\em Proceedings of the IEEE/CVF International Conference on
  Computer Vision (ICCV)}, October 2019.

\bibitem{DFDC}
Brian Dolhansky, Russ Howes, Ben Pflaum, Nicole Baram, and Cristian~Canton
  Ferrer,
\newblock ``The deepfake detection challenge (dfdc) preview dataset,'' 2019.

\bibitem{Celeb-DF}
Yuezun Li, Xin Yang, Pu~Sun, Honggang Qi, and Siwei Lyu,
\newblock ``Celeb-df: A large-scale challenging dataset for deepfake
  forensics,''
\newblock in {\em Proceedings of the IEEE/CVF Conference on Computer Vision and
  Pattern Recognition (CVPR)}, June 2020.

\bibitem{dolhansky2019dee}
B~Dolhansky,
\newblock ``The dee pfake detection challenge (dfdc) pre view dataset,''
\newblock {\em arXiv preprint arXiv:1910.08854}, 2019.

\bibitem{liy2018exposingaicreated}
Chang~M Liy and LYUS InIctuOculi,
\newblock ``Exposingaicreated fakevideosbydetectingeyeblinking,''
\newblock in {\em 2018IEEEInterG national Workshop on Information Forensics and
  Security (WIFS). IEEE}, 2018.

\bibitem{CNN-Aug}
Kaiming He, Xiangyu Zhang, Shaoqing Ren, and Jian Sun,
\newblock ``Deep residual learning for image recognition,''
\newblock in {\em Proceedings of the IEEE conference on computer vision and
  pattern recognition}, 2016, pp. 770--778.

\bibitem{imagenet}
Jia Deng, Wei Dong, Richard Socher, Li-Jia Li, Kai Li, and Li~Fei-Fei,
\newblock ``Imagenet: A large-scale hierarchical image database,''
\newblock in {\em 2009 IEEE conference on computer vision and pattern
  recognition}. Ieee, 2009, pp. 248--255.

\bibitem{sgd}
David Pollard,
\newblock {\em Convergence of stochastic processes},
\newblock Springer Science \& Business Media, 2012.

\bibitem{DeepfakeBench}
Zhiyuan Yan, Yong Zhang, Xinhang Yuan, Siwei Lyu, and Baoyuan Wu,
\newblock ``Deepfakebench: a comprehensive benchmark of deepfake detection,''
\newblock in {\em Proceedings of the 37th International Conference on Neural
  Information Processing Systems}, 2023, pp. 4534--4565.

\bibitem{meso4}
Darius Afchar, Vincent Nozick, Junichi Yamagishi, and Isao Echizen,
\newblock ``Mesonet: a compact facial video forgery detection network,''
\newblock in {\em 2018 IEEE international workshop on information forensics and
  security (WIFS)}. IEEE, 2018, pp. 1--7.

\bibitem{Xception}
Francois Chollet,
\newblock ``Xception: Deep learning with depthwise separable convolutions,''
\newblock in {\em Proceedings of the IEEE Conference on Computer Vision and
  Pattern Recognition (CVPR)}, July 2017.

\bibitem{tan2019efficientnet}
Mingxing Tan and Quoc Le,
\newblock ``Efficientnet: Rethinking model scaling for convolutional neural
  networks,''
\newblock in {\em International conference on machine learning}. PMLR, 2019,
  pp. 6105--6114.

\bibitem{shi2023discrepancy}
Zenan Shi, Haipeng Chen, Long Chen, and Dong Zhang,
\newblock ``Discrepancy-guided reconstruction learning for image forgery
  detection,''
\newblock {\em arXiv preprint arXiv:2304.13349}, 2023.

\bibitem{asam}
Jungmin Kwon, Jeongseop Kim, Hyunseo Park, and In~Kwon Choi,
\newblock ``Asam: Adaptive sharpness-aware minimization for scale-invariant
  learning of deep neural networks,''
\newblock in {\em International Conference on Machine Learning}. PMLR, 2021,
  pp. 5905--5914.

\bibitem{kingma2014adam}
Diederik~P Kingma and Jimmy Ba,
\newblock ``Adam: A method for stochastic optimization,''
\newblock {\em arXiv preprint arXiv:1412.6980}, 2014.

\bibitem{SAGM}
Pengfei Wang, Zhaoxiang Zhang, Zhen Lei, and Lei Zhang,
\newblock ``Sharpness-aware gradient matching for domain generalization,''
\newblock in {\em Proceedings of the IEEE/CVF Conference on Computer Vision and
  Pattern Recognition}, 2023, pp. 3769--3778.

\bibitem{GRADCAM}
Ramprasaath~R Selvaraju, Michael Cogswell, Abhishek Das, Ramakrishna Vedantam,
  Devi Parikh, and Dhruv Batra,
\newblock ``Grad-cam: Visual explanations from deep networks via gradient-based
  localization,''
\newblock in {\em Proceedings of the IEEE international conference on computer
  vision}, 2017, pp. 618--626.

\bibitem{tsne}
Geoffrey~E. Hinton and Sam~T. Roweis,
\newblock ``Stochastic neighbor embedding,''
\newblock in {\em Advances in Neural Information Processing Systems 15 [Neural
  Information Processing Systems, {NIPS} 2002, December 9-14, 2002, Vancouver,
  British Columbia, Canada]}, Suzanna Becker, Sebastian Thrun, and Klaus
  Obermayer, Eds. 2002, pp. 833--840, {MIT} Press.

\bibitem{super-resolution}
Yang He, Ning Yu, Margret Keuper, and Mario Fritz,
\newblock ``Beyond the spectrum: Detecting deepfakes via re-synthesis,''
\newblock in {\em Proceedings of the Thirtieth International Joint Conference
  on Artificial Intelligence}, Aug 2021.

\end{thebibliography}

\end{document}